\documentclass[letterpaper, 10 pt, conference]{ieeeconf}  

\IEEEoverridecommandlockouts                              

\usepackage{bm}
\usepackage{times}
\usepackage{epsfig}
\usepackage{amssymb}
\usepackage{algpseudocode}
\usepackage{algorithm}
\usepackage{amsmath}
\usepackage{subfigure}
\usepackage{graphicx}
\usepackage{multirow}
\usepackage{multicol}
\usepackage{caption}
\usepackage[pdfstartview=FitH,bookmarksnumbered=true,colorlinks,bookmarksopen=true]{hyperref}
\usepackage{mathtools}
\usepackage{booktabs}
\usepackage{pifont}

\DeclarePairedDelimiter\floor{\lfloor}{\rfloor}

\title{\LARGE \bf
GPR: Grasp Pose Refinement Network for Cluttered Scenes
}

\author{Wei Wei\textsuperscript{1, 2, *}, Yongkang Luo\textsuperscript{1, *}, Fuyu Li\textsuperscript{1,2}, Guangyun Xu\textsuperscript{1,2}, Jun Zhong\textsuperscript{1}, Wanyi Li\textsuperscript{1}, Peng Wang\textsuperscript{1, 2, 3,\ding{41}}
\thanks{This work was supported in part by the National Natural Science Foundation of China under Grants (91748131, 62006229 and 61771471), and in part by the Strategic Priority Research Program of Chinese Academy of Science under Grant XDB32050100.}
\thanks{$^{1}$ Institute of Automation, Chinese Academy of Sciences, Beijing, China.} 
\thanks{$^{2}$ School of Artificial Intelligence, University of Chinese Academy of Sciences, Beijing, China}
\thanks{$^{3}$ CAS Center for Excellence in Brain Science and Intelligence Technology, Chinese Academy of Sciences, Shanghai, China}
\thanks{* Authors contributed equally.}
\thanks{\ding{41} Corresponding author:  peng\_wang@ia.ac.cn}
}

\begin{document}

\maketitle
\thispagestyle{empty}
\pagestyle{empty}

\begin{abstract}
Object grasping in cluttered scenes is a widely investigated field of robot manipulation. Most of the current works focus on estimating grasp pose from point clouds based on an efficient single-shot grasp detection network. However, due to the lack of geometry awareness of the local grasping area, it may cause severe collisions and unstable grasp configurations. In this paper, we propose a two-stage grasp pose refinement network which detects grasps globally while fine-tuning low-quality grasps and filtering noisy grasps locally. Furthermore, we extend the 6-DoF grasp with an extra dimension as grasp width which is critical for collisionless grasping in cluttered scenes. It takes a single-view point cloud as input and predicts dense and precise grasp configurations. To enhance the generalization ability, we build a synthetic single-object grasp dataset including 150 commodities of various shapes, and a multi-object cluttered scene dataset including 100k point clouds with robust, dense grasp poses and mask annotations. Experiments conducted on Yumi IRB-1400 Robot demonstrate that the model trained on our dataset performs well in real environments and outperforms previous methods by a large margin.
\end{abstract}

\section{INTRODUCTION}
Robotic grasping is a fundamental problem in the robotics community and has many applications in industry and house-holding service. It has shown promising results in industrial applications, especially for grasping under structured environments, such as automated bin-picking \cite{liu2012fast}. However, it remains an open problem due to the variety of objects in complex scenarios. Objects have different 3D shapes, and their shapes and appearances are affected by lighting conditions, clutter and occlusions between each other.
\begin{figure}
    \centering
    \includegraphics[width=0.9\linewidth]{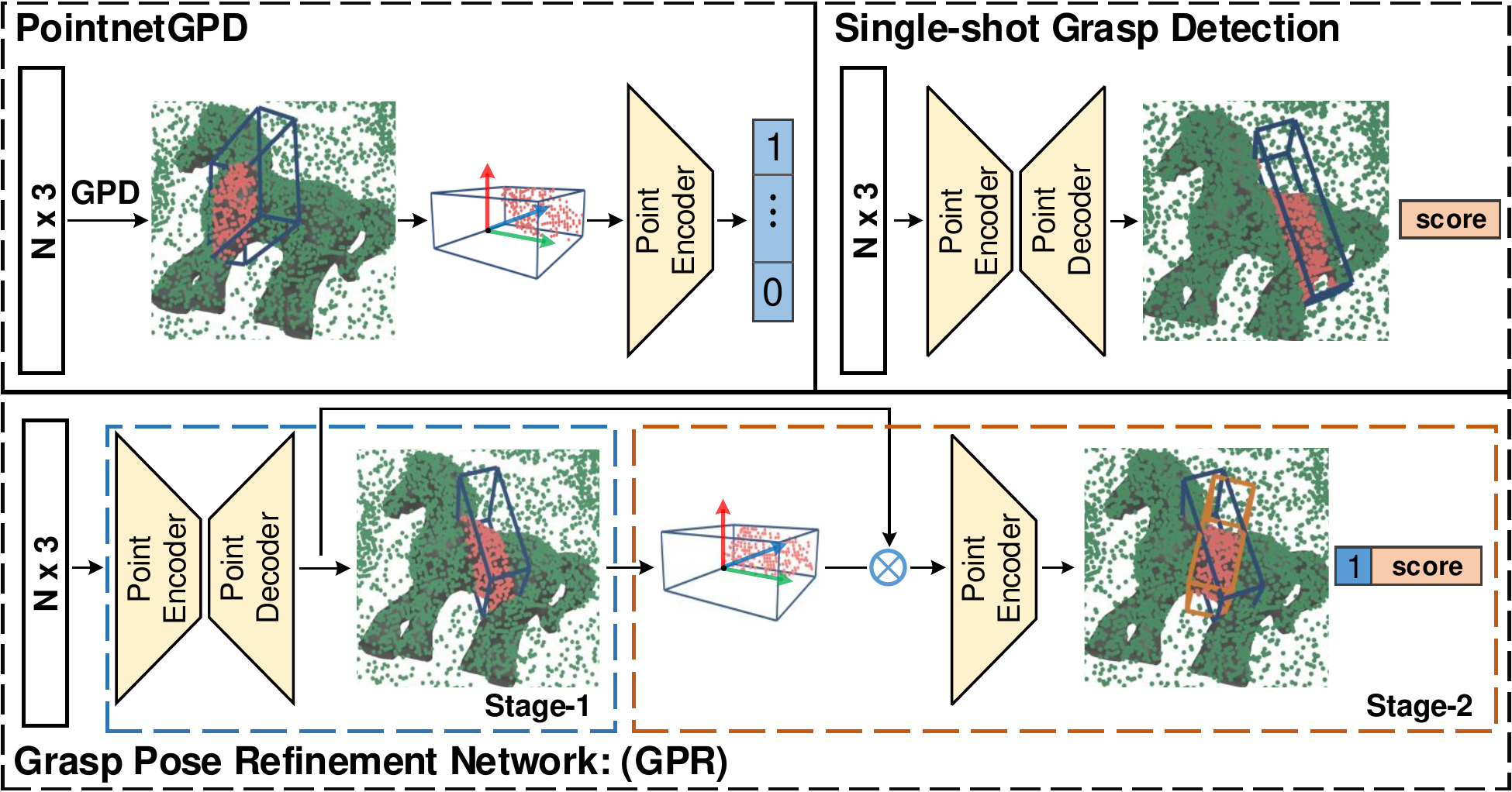}
    \caption{Comparison with state-of-the-art methods. Instead of exhaustive searching and evaluating possible grasp candidates in the point cloud, our method generates possible grasp candidates efficiently in stage-1 as the single-shot grasp detection pipeline. Moreover, our model refines low-quality and classifies noisy grasp candidates in stage-2 base on discriminative feature representation of the local grasping area.}
    \label{fig:idea}
    \vspace{-6mm}
\end{figure}

Traditionally, the problem of object grasping in cluttered scenes is tackled by estimating 6D object pose \cite{zeng2017multi,posecnn,wang2019densefusion} and selecting grasp from the grasp database. As a result, these approaches are not applicable to unseen objects. In order to generalize to unseen objects, many recent works \cite{image_grasp,image_grasp_multi_obj,roi_grasp,dexnet2.0,MorrisonLC18} conduct grasp pose detection as  rectangle detection in 2D space with CNNs, and their models perform well on novel objects. However, planar grasping with 3/4 DoF (degree of freedom) inevitably results in inflexibility, since the gripper is forced to approach objects vertically. Besides the DoF constraint, these works utilize the 2D images as input, which ignore gripper contact with the object in 3D space. Some recent works suggest that 3D geometry structure is highly relevant to grasp quality \cite{gpd,pointnetgpd}. PointNetGPD \cite{pointnetgpd} evaluates grasp quality in 3D space with exhaustive searching in point clouds. S4G \cite{s4g} and PointNet++Grasping \cite{pointnet++_grasping} propose efficient single-shot grasp pose detection network architectures, while the results may be noisy and suffer collisions with surrounding objects. The main reason can be attributed to: 1) lack of shape awareness of the local contextual geometry of the gripper closing area; 2) grasping with max opening width is more likely to cause collisions with surrounding objects in dense clutter.      

Considering the above problems, we propose to detect grasp poses globally and refine them locally. Single-shot feature representation helps to avoid exhaustive searching in the point cloud, while it is not able to learn discriminative local feature representation without further inspection of the local grasping area. For addressing the limitation, we turn to focus on the local grasping area and design a two-stage grasp pose refinement network (\textbf{GPR}) for estimating stable and collisionless grasps from point clouds. As illustrated in Fig.\ref{fig:idea}, our model predicts coarse and noisy grasp proposals in the first stage. Then, points inside the proposals are cropped out and transformed into local gripper coordinate in the second stage. Finally, these points are used to encode discriminative local feature representation for grasp proposals refinement and classification. Remarkably, our model takes a single-view point cloud as input and extends the 6-DoF grasp with an extra dimension as grasp width, which adjust gripper opening width and avoid unnecessary collisions. Furthermore, our two-stage network is trained in an end-to-end fashion.

For most data-driven methods, it is common to boost generalization performance with a large-scale dataset. However, manually annotated 6D grasps can be time-consuming \cite{grasp_1_billion}. Most current works generate grasp annotations  based on traditional analysis methods \cite{force_closure, ferri_canny} or physics simulators \cite{graspit,pybullet,mujoco}. In \cite{pointnetgpd,dexnet2.0}, researchers had built datasets for individual objects, while ignoring multi-objects in cluttered scenes. \cite{s4g,pointnet++_grasping} propose to generate grasps in cluttered scenes. However, almost all of the object models come from the YCB object dataset \cite{ycb_video} may lead to insufficient shape coverage. We collect 150 objects with various shapes and build large-scale synthetic datasets for both individual objects and objects in dense clutter. Experiment results show that the model trained on our dataset performs well in the real robot platform and gets promising results.

In summary, our primary contributions are:
\begin{itemize}
   \item An end-to-end grasp pose refinement network for high-quality grasp pose prediction in cluttered scenes that detects globally while refines locally.
   \item Extend 6-DoF grasp with grasp width as a 7-DoF grasp for improvement of dexterous and collisionless grasping in dense clutter.
   \item  A densely annotated synthetic single-object grasp dataset including 150 object models, and a large scale cluttered multi-object dataset with 100k point clouds with detailed annotations. We will release the dataset.
\end{itemize}

\section{RELATED WORK}
\textbf{Deep Learning based Grasp Configuration Detection.} \cite{caldera2018review} gives a thorough survey of robotic grasping based on deep learning. Given the object model and grasp annotations, \cite{collet2011moped,zeng2017multi} tackle this problem as template matching, and the 6-DoF pose retrieving problem. While template matching methods show low generalization ability for unknown objects. \cite{gpd} designs several projection features as the input of a CNN-based grasp quality evaluation model. \cite{pointnetgpd} replaces input with direct irregular point cloud and train PointNet \cite{pointnet} for grasp classification. These methods rely on detailed local geometry for constructing both collision-free and force-closure grasps. \cite{image_grasp, image_grasp_multi_obj, roi_grasp,image_grasp2,image_grasp3} tackle this problem as grasp rectangle detection in 2D images from a single object to multi-object scenarios. While these methods just perform 3/4-DoF grasp. \cite{s4g} proposes a single-shot grasp proposal framework to regress 6-DoF grasp configurations from point cloud directly. \cite{pointnet++_grasping} follows a similar setting, while it generates grasp based on the assumption that the approaching direction of a grasp is along the surface normal of the objects. Worth noting that \cite{dexnet2.0} collects numerous object models for GQ-CNN training and obtains state-of-the-art performance. Of all the above methods, GPD can also estimate grasp width with geometry prior. However, it relies on multi-view point clouds input. In this paper, we revisit grasp width as a critical element for grasp configuration and our model can directly predict high accuracy grasp width.

\textbf{Grasping Dataset Synthesis.}   \cite{cornell_grasp_dataset,image_grasp_multi_obj,roi_grasp} annotate rectangle representation for grasping detection in images manually.  \cite{pinto2016supersizing,hand_eye_coordinate} collect annotations with a real robot. While an enormous amount of annotated data is needed for supervised deep learning, therefore manually grasp configuration annotating is unpractical due to time-consuming. Given an object with a gripper model and environment constraints, we are able to synthesize grasp configurations in two kinds of ways generally. One is based on analytic methods \cite{bicchi2000robotic}, which derive from force-closure \cite{force_closure} and Ferrari Canny metric \cite{ferri_canny}. \cite{grasp_survey} gives a detailed survey of these methods. \cite{s4g,pointnet++_grasping, pointnetgpd,dexnet1.0,dexnet2.0,grasp_1_billion} generate dataset based on this way. Another is based on physical simulators, such as \cite{pybullet,mujoco}, these simulators perform better than analytic methods in terms of force contacts.  \cite{6-dof_graspnet,depierre2018jacquard,contact_grasp,yan2018learning,yan2019data} generate their dataset using simulated environment.

\textbf{Deep Learning on Point Cloud Data.} PointNet \cite{pointnet} and PointNet++ \cite{pointnet++} are two novel frameworks to directly extract feature representation from point cloud data. Many methods \cite{pointconv, kpconv,pointcnn,densepoint,pointrcnn,votenet} extend these frameworks to point cloud classification, detection and segmentation. In this paper, we utilize PointNet++ as the backbone.

\section{PROBLEM STATEMENT}

In this work, we focus on the problem of planning a robust two-fingered parallel-jaw grasping based on point clouds. Our two-stage refinement network takes the whole cluttered scene as input and outputs dense grasp poses with high quality and robustness. Some of the key definitions are introduced here: 

\textbf{Object States:} Let $\bm {x}_i=\left ( \mathcal{O}_i, \bm T_i, \bm \gamma \right )$ describes state of an object in a grasp scene, where $\mathcal{O}_i$ specifies the surface model, mass and centroid  properties of object $i$, $\bm T_i$ denotes 6D object pose, $\bm \gamma$ denotes friction coefficient. 

\textbf{Point Clouds:} Let $\bm {y}_k \in \mathbb{R}^{N\times3}$ represents the point cloud of the $k^{th}$ scene captured by the depth camera.

\textbf{Grasps:} Let $\bm G=\left [ \bm g_1, \bm g_2, \bm \cdots \bm g_m \right ]$ denotes grasp configurations in a cluttered scene. Each grasp configuration is defined as $\bm g_i=\left ( \bm o, \bm n, \bm r, \omega,  c_1, c_2\right )$, where $\bm o=\left ( o_x, o_y, o_z \right ) $ represents the origin lies at the middle of the line segment connecting two finger tips, $\bm n=\left ( n_x, n_y, n_z \right )$ and $\bm r=\left ( r_x, r_y, r_z \right )$ denote approach direction and closing direction of a grasp, $w$ describes grasp width, $c_1$ and $c_2$ denote contact points. 

\textbf{Grasp Metric:} We adopt the widely used Ferrari Canny metric \cite{ferri_canny} for labelling grasp quality.

\section{DATASET GENERATION}
In this section, we introduce our dataset generation method for grasp poses annotation for both individual objects and objects in dense clusters. The overall pipeline is illustrated in Fig.\ref{fig:dataset_pipeline}. We take the following procedure to obtain dense grasp annotations. Firstly, we label single-object grasp annotations and then match grasp annotations into cluttered scenes according to the 6D object pose. Finally, we apply collision filtering for all the grasp configurations.

\subsection{Single-object grasp dataset Generation}
For single-object grasp dataset generation, we collect 150 objects of various shapes and categories. Half of these objects come from the BOP-Challenge dataset and YCB-Video dataset \cite{ycb_video}, others are collected from the internet. 

Given a specific object model $\mathcal{O}$, the target is to generate dense grasp annotations including grasp configuration $g$ and corresponding grasp metric mentioned above.
\begin{figure}
    \centering
    \includegraphics[width=0.75\linewidth]{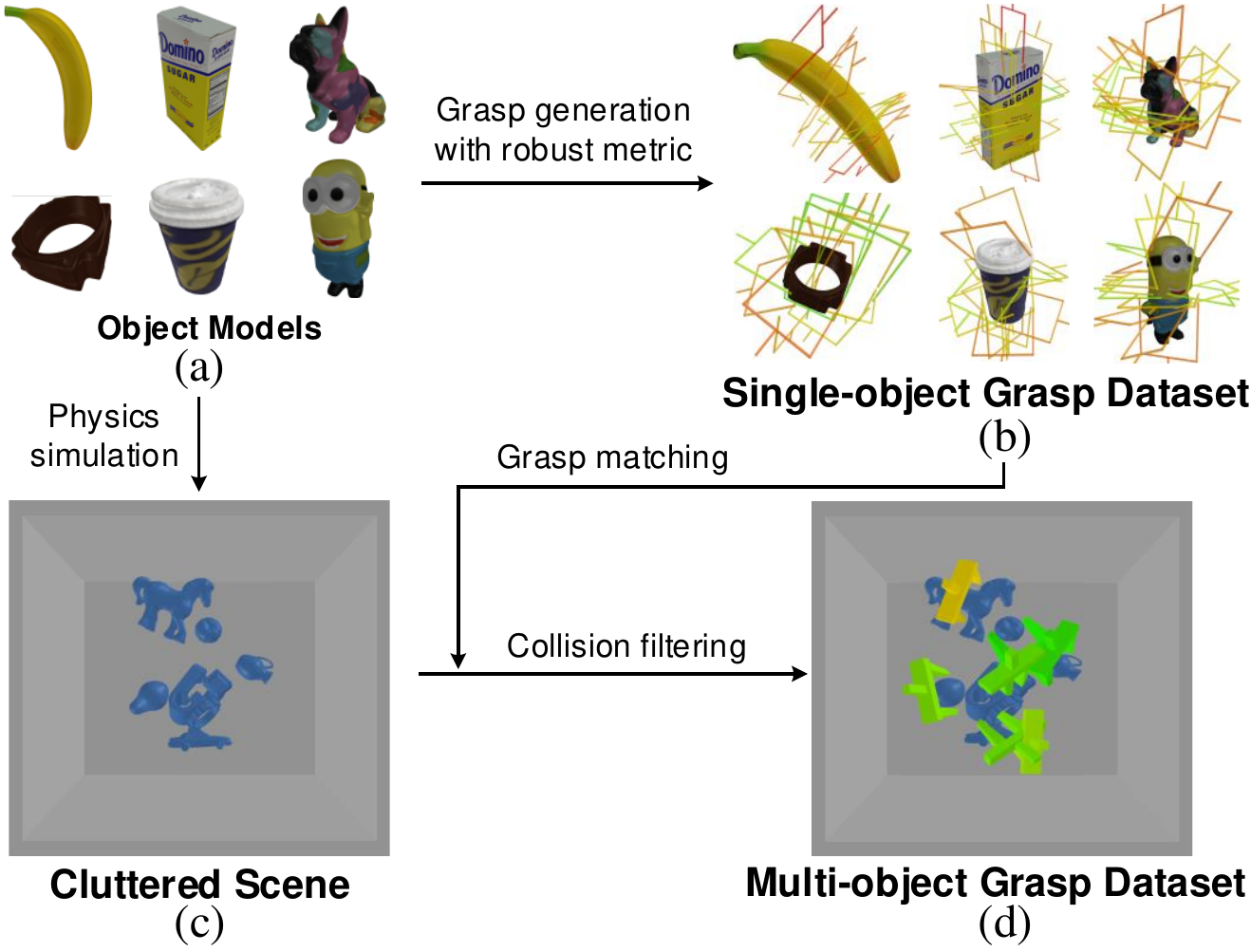}
    \caption{Overview of our datasets generation procedure. (a) Example single object models. (b) Example single object grasps with label $\bm Q({\bm g})$. For each object, 15 grasps are sampled for visualization. The colors from red to green represent $\bm Q({\bm g})$ from low to high. (c) Illustration of a cluttered scene. (d) Example grasps in a cluttered scene.}
    \label{fig:dataset_pipeline}
    \vspace{-6mm}
\end{figure}
First, $\bm N$ candidate contact points $\bm p_1, \bm p_2, \bm \cdots \bm p_N$ are sampled on object surface model with outward normals $\bm n$ calculated. Based on force-closure principle, $\bm k$ antipodal grasp directions are then sampled inside the friction cone of point $\bm p_i$. Each antipodal grasp candidate $\bm g_i$ will be classified as a positive grasp candidate $\bm g_i^T$, if satisfies rules as follows: 1) At least one antipodal contact point $\bm c_2$ is found on object backward surface; 2) Force-closure property. Otherwise,  antipodal grasp candidate is classified as negative grasp $\bm g_i^F$. 

Second, for each positive antipodal grasp candidate $\bm g_i^T$ of a contact point $\bm p_i^T$, collision check is applied between gripper and object. Those grasp candidates failed in collision check will be classified as negative grasps $\bm g_i^F$. If no positive antipodal grasp candidate is reserved, the corresponding sampled point is classified as a negative point $\bm p_i^F$, which means an unsuitable contact point.

Third, the grasp metric for each reserved positive grasp candidates is calculated by Ferrari Canny metric as $\bm Q({\bm g})$.

Finally, we apply  Non-maximum Suppression algorithm (NMS) for pruning redundant grasps. Distance between two sampled grasp $g_1$ and $g_2$ is calculated by following equation:
\begin{small}
\begin{equation}
\begin{aligned}
     \bm D(g_1, g_2)&=\beta_1 \cdot \bm{||} ((c_1+c_2)/2|{g_1}) - ((c_1+c_2)/2|{g_2}) \bm{||}_2 \\
     &+ \beta_2 \cdot arccos(\bm{|}(\bm{\gamma}|g_1) \cdot (\bm{\gamma}|g_2)\bm{|})/\pi \\
     &+ \beta_3 \cdot arccos((\bm{n}|g_1) \cdot (\bm{n}|g_2))/\pi.
\end{aligned}
\label{equation:grasp_distance}
\end{equation}
\end{small}
$\bm n, \bm{\gamma}$ are set to 16384, 0.3. $\bm {\beta_1}, \bm {\beta_2}$, and $\bm {\beta_3}$ is set to 1, 0.03, and 0.03 in our experiments. For all the objects $\mathcal{O}_1, \mathcal{O}_2, \cdots \mathcal{O}_n$, the output annotations are denoted as $\{\bm g_i, \bm Q(\bm g_i) \ | \ \mathcal{O}_1, \mathcal{O}_2, \cdots \mathcal{O}_n \}$. Examples of our single-object grasp dataset are shown in Fig.\ref{fig:dataset_pipeline} (b).
\subsection{Multi-object Grasp Dataset Generation} 
To simulate densely cluttered scenes for the multi-object grasp dataset, we adopt the following procedures using \cite{pybullet}:

First, $m$ objects are randomly sampled, then these sampled objects are initializing with random poses, and falling into a static bin successively in the simulator, as shown in Fig.\ref{fig:dataset_pipeline}(c). 

Then, the 6D object pose will be recorded after all sampled objects falling into the bin and reaching stable states. Each unsuitable grasp point $p_i^F$ for each object $\mathcal{O}_i$ will be added into negative point set $p_{neg}$. Then we apply collision check for each grasp $g_j$ of each object $\mathcal{O}_i$ obtained by single-object grasp generation. If no collision occurs, contact points $c_1, c_2$ of grasp $g_i$ will be added into positive grasp contact points set $p_{pos}$, and the corresponding grasp annotation will be added into positive grasp set $g_{pos}$. Otherwise, the point will be added into negative grasp contact points set $p_{neg}$.

\begin{figure}
    \centering
    \includegraphics[width=0.8\linewidth,height=0.45\linewidth]{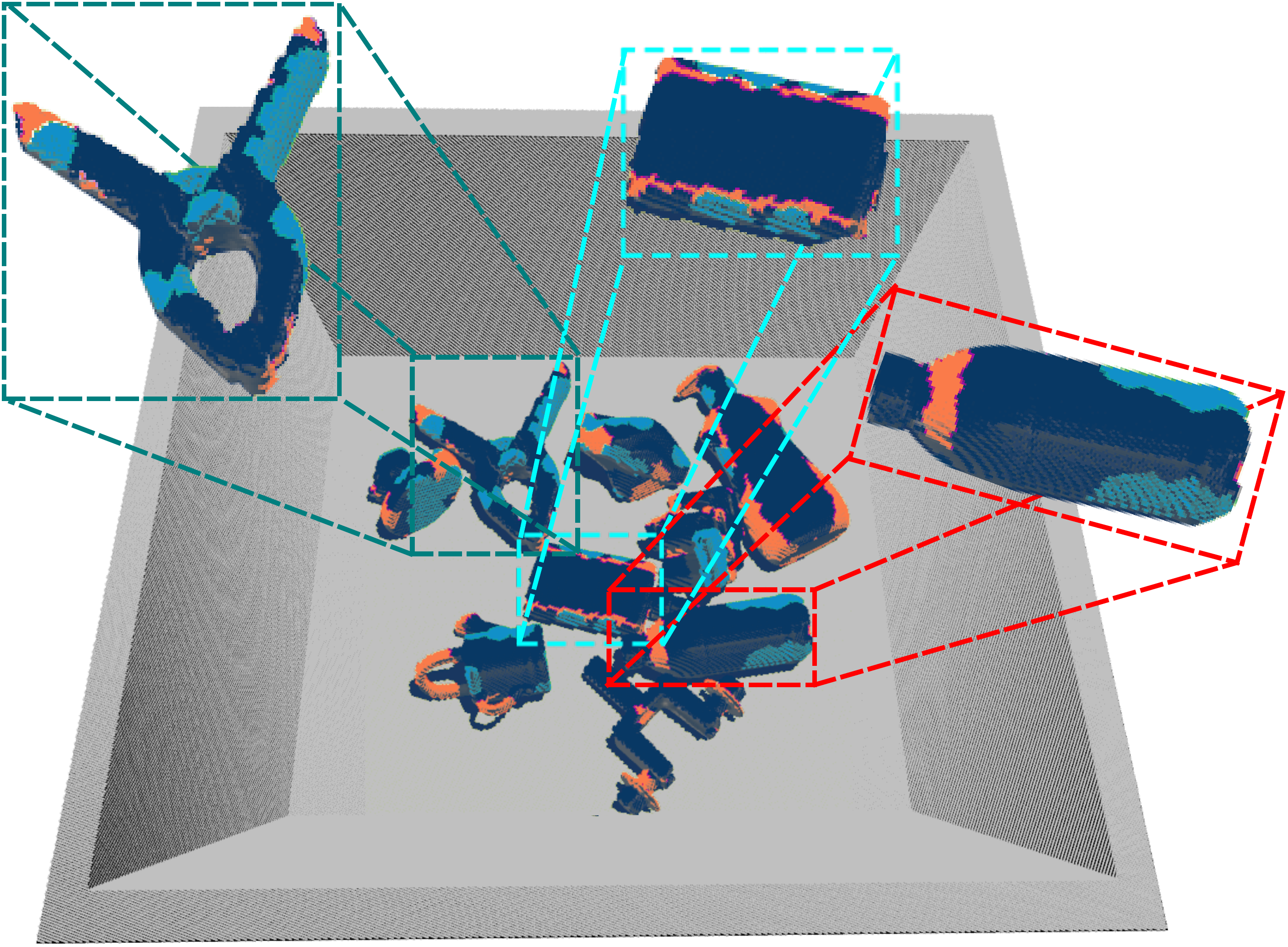}
    \caption{An example shows points mask label $M(p)$ in our multi-object grasp dataset. Points in light blue denote positive grasp contact point. Points in dark blue denote negative contact point due to collision. Points in orange denote unsuitable contact points on foreground objects.}
    \label{fig:multi_obj_mask}
    \vspace{-6mm}
\end{figure}

Point cloud $y_k$ within the bin is cropped for generating points label and mask which is defined as follows:
\begin{equation}
\begin{aligned}
    L(p_i) &= [\bm n, \bm \gamma, \omega, Q(g_i)], \\
    M(p_i) &= [\mathbb{I}\big(Q(g_i)\big)], \\
    \mathbb{I}\big(Q(g_i)\big)&= 
        \begin{cases}
         1 & \text{if } Q(g_i) > 0, \\
    0  & \text{otherwise}. 
\end{cases}
\end{aligned}
\end{equation}
Where $\mathbb{I}$ denotes Indicator function for generating points mask. For each point $p_i \in p_{pos}$, a KD-Tree search is applied to find the nearby points $p_i^R=[p_{i_1}, p_{i_2}, \cdots p_{i_k} | \bm p_i, y_k, R]$ among $y_k$ with query radius $R$. Moreover, each point in $p_i^R$ will be broadcast with the same label $L(p_i)$ and mask $M(p_i)$. Finally, each point will only reserve the corresponding label and mask with the highest score. For each point $p_i \in p_{neg}$, the similar process will be done. An Example is shown in Fig.\ref{fig:multi_obj_mask}.
\begin{figure*}
    \begin{center}
		\includegraphics[width=0.85\linewidth,height=0.275\linewidth]{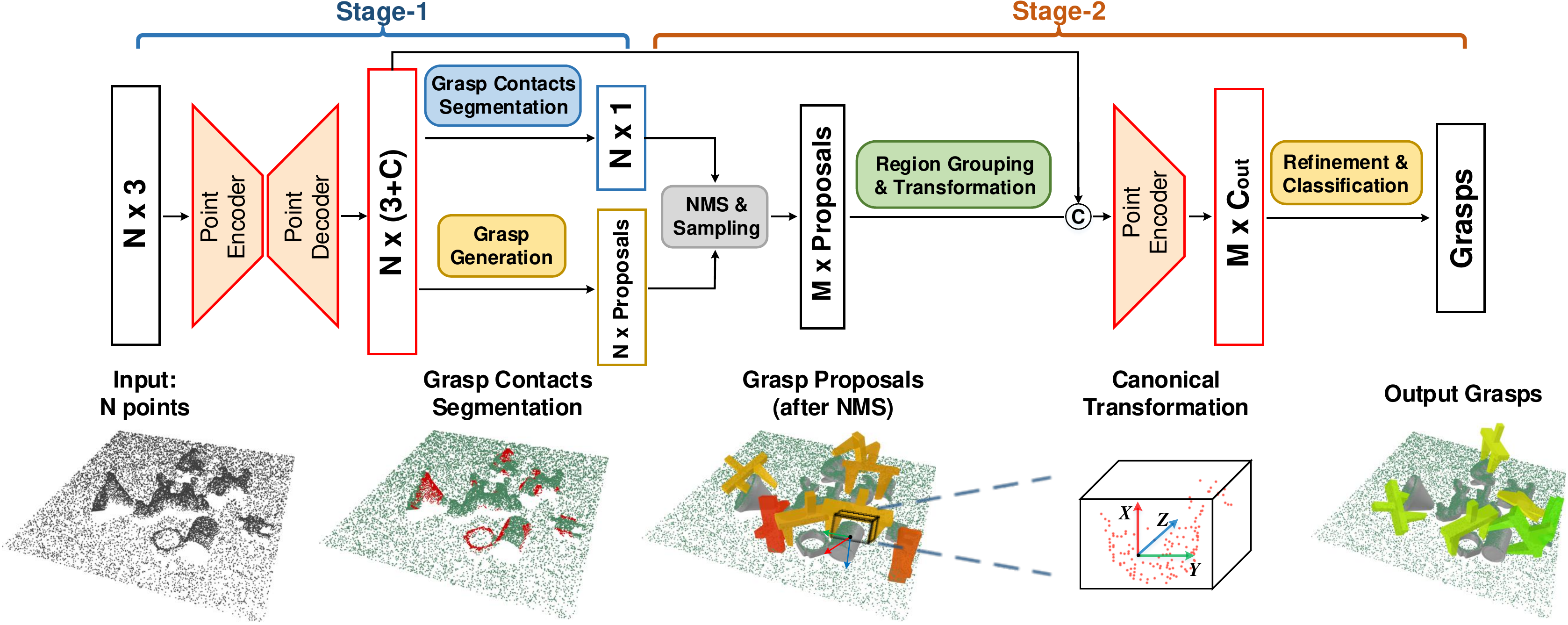}
	\end{center}
    \caption{Overview of our \textbf{GPR} network for grasp pose detection and refinement in point cloud. Stage-1 for generating 7-DoF grasp proposals. Stage-2 for refining grasp proposals with further geometry awareness of local grasping area.}
    \label{fig:pipeline}
    \vspace{-6mm}
\end{figure*}

\section{GRASP POSE REFINEMENT NETWORK}
In this section, we present our proposed two-stage grasp pose refinement network (\textbf{GPR}) for grasp pose detection in cluttered scenes. The overall structure is illustrated in Fig.\ref{fig:pipeline}.

\subsection{Grasp Proposal Generation}
Existing 6-DoF grasp pose detection methods could be classified into one-stage and two-stage methods. One-stage methods\cite{image_grasp2,image_grasp3,dexnet1.0,dexnet2.0,s4g,pointnet++_grasping} are generally faster but directly predict grasp pose without local geometry awareness. Two-stage methods\cite{image_grasp,image_grasp_multi_obj,roi_grasp} mostly depend on anchor mechanism\cite{faster_rcnn} developed on 2D object detection, which generate proposals firstly and then refine the proposals and confidences in the second stage. However, directly applying anchor mechanism for predicting grasp pose in 3D space is non-trivial due to the huge search space and irregular format of the point cloud.

Therefore, we propose to directly estimate grasp pose in a bottom-up manner to avoid exhaustive searching in 3D space with 3D rotation inspired by \cite{s4g,pointnet++_grasping}. We predict mask and coarse 7-DoF grasp proposal for each point in the scene, as shown in stage-1 sub-network of Fig.\ref{fig:pipeline}.

\textbf{Feature representations and segmentation.} We design the backbone network based on the PointNet++ \cite{pointnet++}, which is a robust learning model for dealing with sparse point cloud and non-uniform point density. We utilize the PointNet++ network with multi-scale grouping strategy as the backbone. 

Given the point-wise feature encoded by the backbone network, we append two head ahead to our backbone: one segmentation head for predicting grasp contact points mask, and one grasp pose regression head for generating 7-DoF grasp proposals. We utilize focal loss \cite{focal_loss} to handle the severe imbalance problem for grasp contacts segmentation, as shown in Fig.\ref{fig:pipeline}. 

\begin{figure}
    \centering
    \subfigure[]{\includegraphics[width=0.35\linewidth]{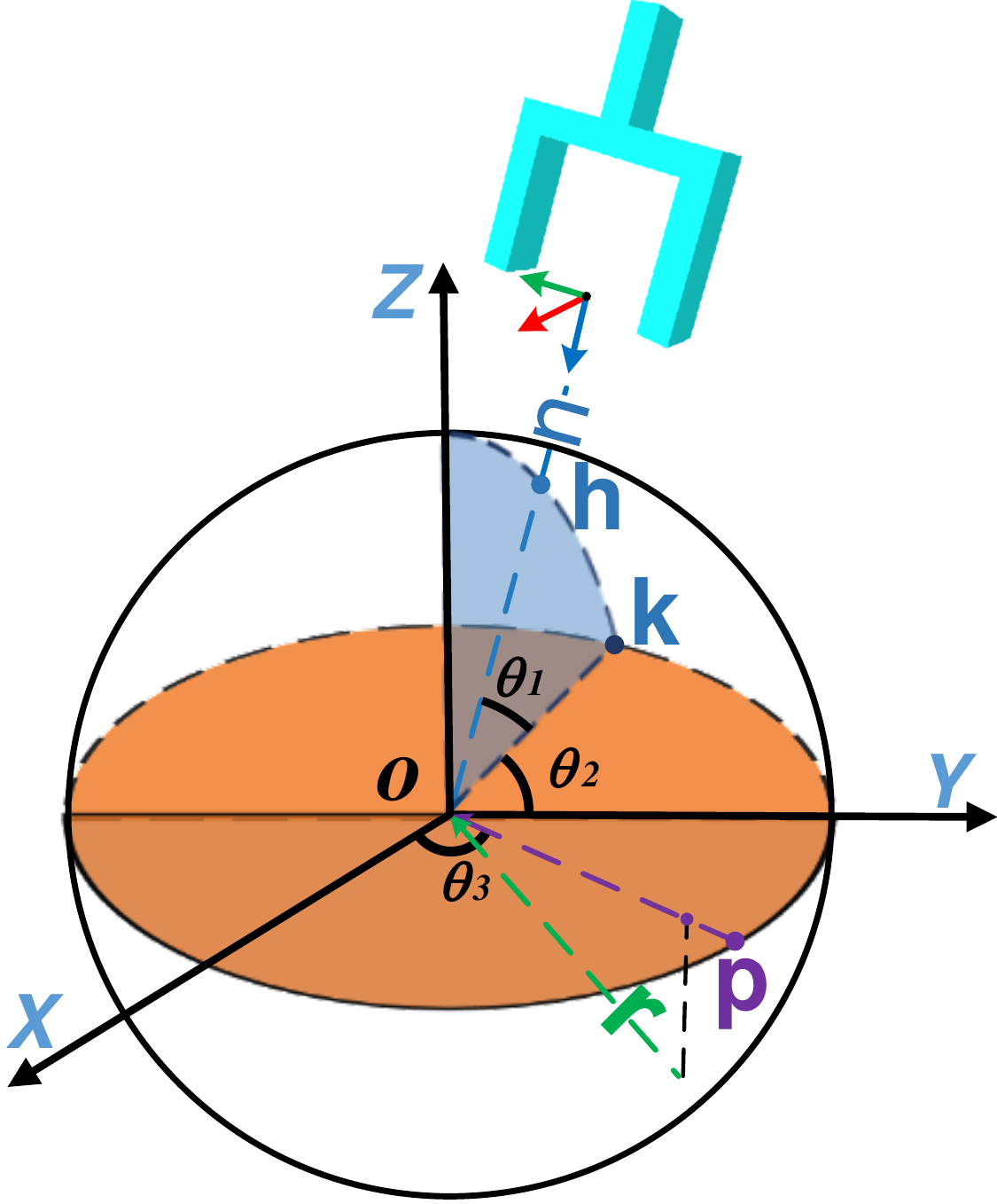}
    \label{fig:bin_base_grasp_a}}
    \subfigure[]{\includegraphics[width=0.35\linewidth]{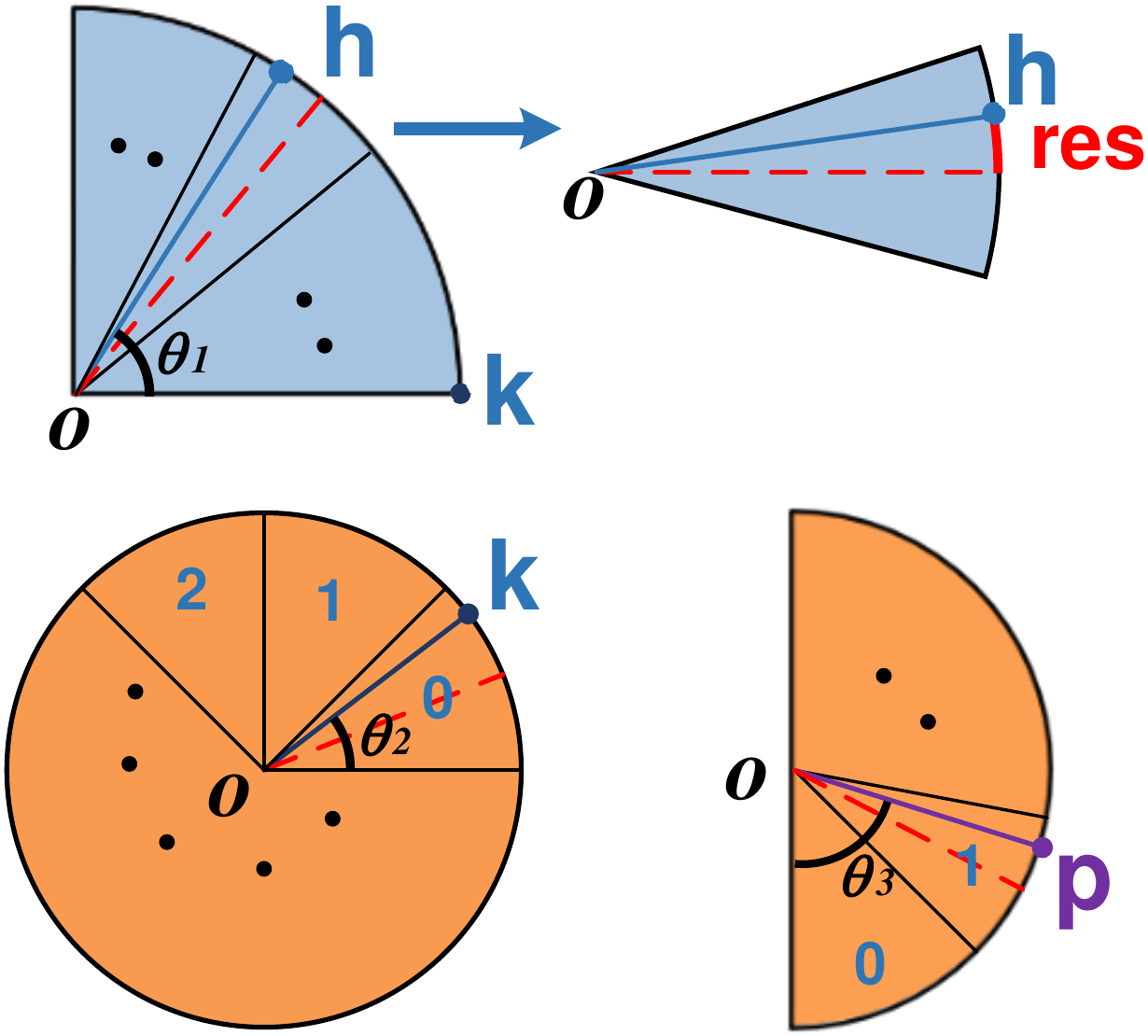}
    \label{fig:bin_base_grasp_b}}
    \caption{An illustration of bin-based angle regression. \subref{fig:bin_base_grasp_a} The grasp approach vector $\textbf{n}$ is denoted by azimuth angle \textbf{$\theta_1$} and elevation angle \textbf{$\theta_2$}, closing vector \textbf{r} is projected to X-Y plane and denoted by azimuth angle \textbf{$\theta_3$}. \subref{fig:bin_base_grasp_b} Examples show range of azimuth and elevation angle are split into a series of bins, where \textbf{res} denotes normalized residual value within the bin.}
    \label{fig:bin_base_grasp}
    \vspace{-4mm}
\end{figure}

\textbf{Bin-based grasp pose regression.} It is difficult to regress 7-DoF grasp configuration directly, which has been proved in previous literature \cite{ssd_6d,pointrcnn,votenet}. Therefore, we develop bin-based regression method similar as \cite{pointrcnn}. Specifically, a 7-DoF grasp is represented as $\bm g = ( \bm o, \bm n, \bm r, \omega)$, where $\bm o=(x, y, z)$ denotes the grasp center, $\bm n$ and $\bm r$ denote approach and closing directions of the gripper, $\omega$ denotes gripper opening width. Gripper direction regression is converted to angle prediction, as show in Fig.\ref{fig:bin_base_grasp}. For angle prediction, gripper approach vector is denoted by $\theta_{1} \in [0, 2\pi]$ and $\theta_{2} \in [0, \pi/2]$ jointly, while finger closing direction is projected onto X-Y plane, and denoted by $\theta_{3} \in [-\pi/2, \pi/2]$. 

We divide a target angle of point $p$, \textit{e.g.} $\theta_1^p$, into $n$ bins with uniform angle $\delta_{\theta_1}$, and calculate the bin classification target bin$_{\theta_1}^{p}$ and residual regression target res$_{\theta_1}^{p}$ within the classified bin. The angle loss for $\theta_1$, $\theta_2$ and $\theta_3$ consists of two terms, one term for bin classification and another for residual regression within the classified bin. The target angle could be formulated as follows:
\begin{equation}
\begin{aligned}
    \mathop{\text{bin}_{\theta}^{p}}\limits_{\theta \in \{\theta_{1,2,3}\}}  &= \floor*{\frac{\theta^p - \theta_{s}}{\delta_{\theta}}}, \\
    \mathop{\text{res}_\theta^{p}}\limits_{\theta \in \{\theta_{1, 2, 3}\}}  &= \frac{1}{\delta_{\theta}}\left( \theta^p - \theta_{s} - \left(\text{bin}_\theta^{p} \cdot \delta_{\theta} + \frac{\delta_{\theta}}{2}\right)\right).
\end{aligned}
\end{equation}
Where $\theta^{p}$ ($\theta \in \{\theta_1, \theta_2, \theta_3\}$) is the target grasp angle of a specific grasp contact point $p$, $\theta_{s}$ denote the starting angle, bin$_{\theta}^{p}$  is the ground-truth bin assignment, res$_{\theta}^{p}$ is the residual value for further angle regression within the assigned bin, and $\delta_{\theta}$ is the unit bin angle of $\theta$ for normalization.

For grasp center and grasp width prediction, we adopt the following formulation:
\begin{small}
\begin{equation}
\begin{aligned}
\mathop{\text{bin}_u^{p}}\limits_{u \in \{x, y, z, \omega\}}  &= \floor*{\frac{u^p - u^{p_c} + \mathcal{S}_u}{d_u}}, \\
\mathop{\text{res}_u^{p}}\limits_{u \in \{x, y, z, \omega\}}  &= \frac{1}{d_u}\left( u^p - u^{p_c} + \mathcal{S}_u - \left(\text{bin}_u^{p} \cdot d_u + \frac{d_u}{2}\right)\right).
\end{aligned}
\end{equation}
\end{small}
Where $(x^p, y^p, z^p)$ is the coordinates of an interest grasp contact point, $\omega^p$ is the grasp width, ($x^{p_c}, y^{p_c}, z^{p_c}$) and $\omega^{p_c}$ is the grasp center coordinates and grasp width of its corresponding grasp configuration. The $\text{bin}_u^{p}$ and $\text{res}_u^{p}$ $(u \in \{x, y, z, \omega\})$ are ground-truth bin assignment and residual location within the assigned bin, and $d_u$ is the bin length for normalization. $\mathcal{S}_u$ denotes the corresponding search range.

The overall loss of grasp proposal generation sub-network could be formulated as follows:
\begin{equation}
\begin{aligned}
    \mathcal{L}_{\text{grasp}}^p &= \sum_{u \in \{x,y,z,\omega,\theta_{1,2,3}\}} ( \mathcal{F}_\text{cls}(\widehat{\text{bin}}_u^{p}, \text{bin}_u^{p}) \\
    & \qquad + \mathcal{F}_\text{reg}(\widehat{\text{res}}_u^{p}, \text{res}_u^{p}) ), \\
    \mathcal{L}_\text{stage-1} &=\! \frac{1}{N_{\text{pos}}}\sum_{p \in \text{pos}}\mathcal{L}_\text{grasp}^{p} + \sum_{}\mathcal{L}_{\textrm{focal}}^p(y_t).
\end{aligned}
\end{equation}
The loss $\mathcal{L}_\text{stage-1}$ includes two terms,  $\mathcal{L}_{\text{grasp}}$ for grasp poses prediction and $\mathcal{L}_{\text{focal}}$ for grasp contact points segmentation. Where   $N_{\text{pos}}$ is the number of positive grasp contact points, $y_t$ is the probability of point $p$ as a positive grasp contact point. Where $\widehat{\text{bin}}_u^{p}$ and $\widehat{\text{res}}_u^{p}$ are the predicted bin assignment and residual of point p, ${\text{bin}}_u^{p}$ and ${\text{res}}_u^{p}$ are corresponding ground-truth. $\mathcal{F}_\text{cls}$ denotes the classification loss of bin assignment, and $\mathcal{F}_\text{reg}$ denotes regression loss for residual prediction. 

\subsection{Grasp Proposal Refinement}
\textbf{Non-maximum suppression and Grasp proposal sampling.} Since sub-network for stage-1 generates one proposal per point, there are a larger number of proposals around ground-truth grasps. Non-maximum suppression (NMS) is applied to select the local maximum. 

\textbf{Region grouping and grasp canonical transformation.} Given the grasp proposals generated by stage-1, point clouds within the gripper closing area are cropped out for further feature representation learning. Unified local coordinates are utilized to eliminate the ambiguity caused by absolute coordinate for objects with various poses and locations. Specifically, we adopt canonical transformation for points within the gripper closing area as shown in Fig.\ref{fig:pipeline}. We set Approaching, Closing, and Orthogonal directions of the gripper as X, Y, and Z axes respectively, and the origin locates at the gripper bottom center. In experiments, the gripper closing area is enlarged by a scalar $\epsilon$ to capture more contextual information, which helps for proposal refinement.  

\textbf{Feature learning for grasp proposal refinement.} After proposal canonical transformation, fine-grained local features within the proposals will be learned with the following steps.

First, for each point within the enlarged 3D grasp proposal, we obtain its canonical coordinate $\tilde{p}$ = $\mathcal{T}(p)$ =  $(x^{\tilde{p}}, y^{\tilde{p}}, z^{\tilde{p}})$ and corresponding global semantic feature learned by stage-1. Then, each inside point $\tilde{p}$ and corresponding feature $f^{p}$ of each grasp proposal are combined. Finally, the concated feature of each point inside the proposal are fed into a point cloud encoder to fuse both the global and local feature. Thus, we can obtain discriminative feature representation for grasp proposal refinement with grasp width and confidence.

The overall loss for training grasp proposal refinement sub-network is similar as depicted in grasp proposal generation sub-network.

\section{EXPERIMENTS}
We evaluate our GPR network both in simulation and the Yumi IRB-1400 Robot platform. In simulation experiments, ablation studies show our model predicts high precision grasp configurations. In the real robot platform, experimental results show that our model has good generalization ability.
\subsection{Implementation Details}
For each point cloud grasp scene, 16384 points are sampled as input. The learning rate
is set to 0.02 at start, and it is divided by 10 when the error plateaus. During the training phase, 256 proposals are sampled after proposals NMS for stage-2, while 100 proposals for inference. Of all the 150 object models, 120 objects are selected for training. Of all the 100k point clouds, 80k point clouds as training data. 
\subsection{Simulation Experiments}
\subsubsection{Extend 6-DoF Grasp with Grasp Width}
\begin{figure}
    \centering
    \subfigure[]{\includegraphics[width=0.625\linewidth,height=0.3\linewidth]{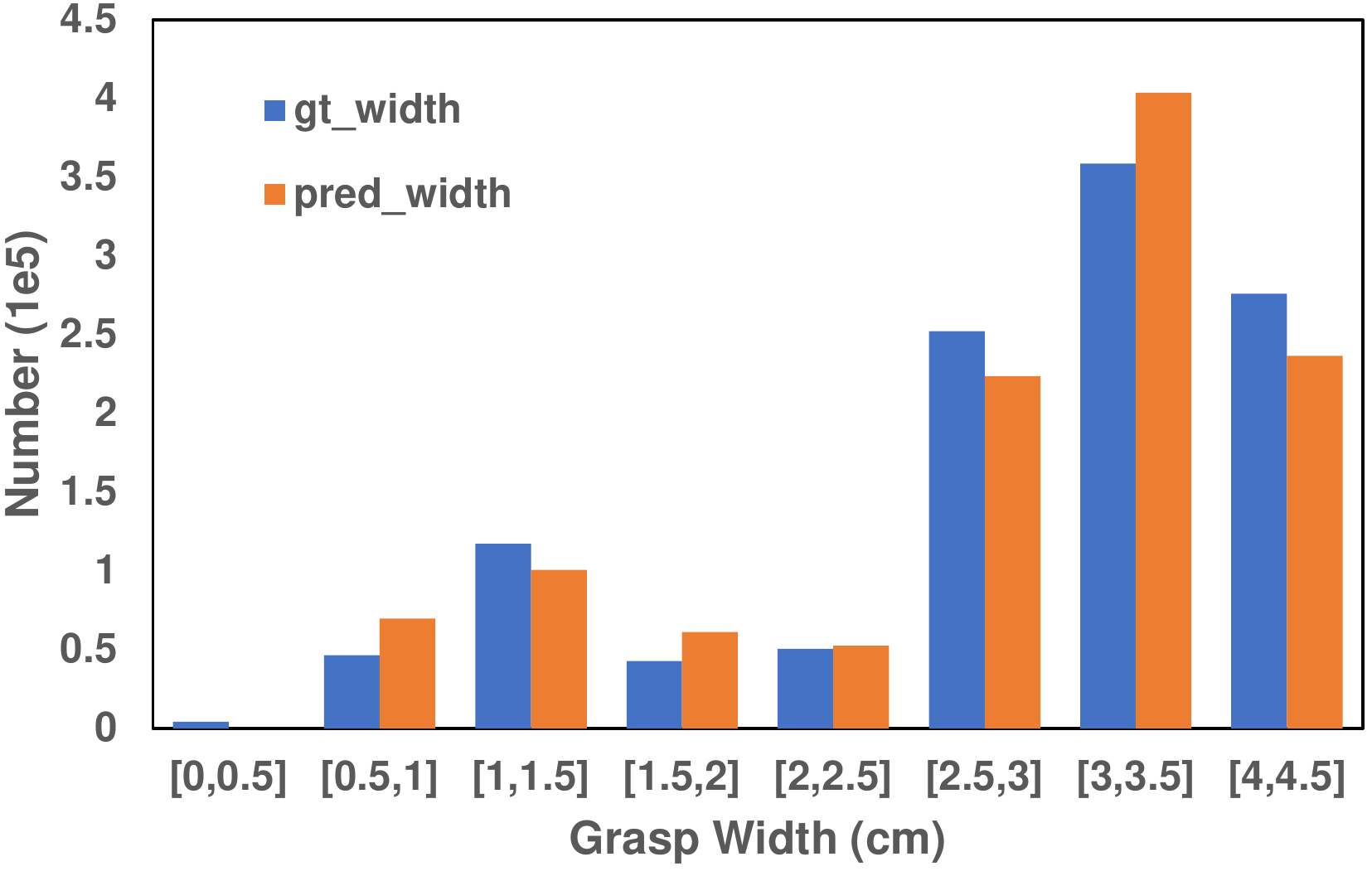}
        \label{fig:grasp_width_distribution}}
    \subfigure[]{\includegraphics[width=0.325\linewidth,height=0.3\linewidth]{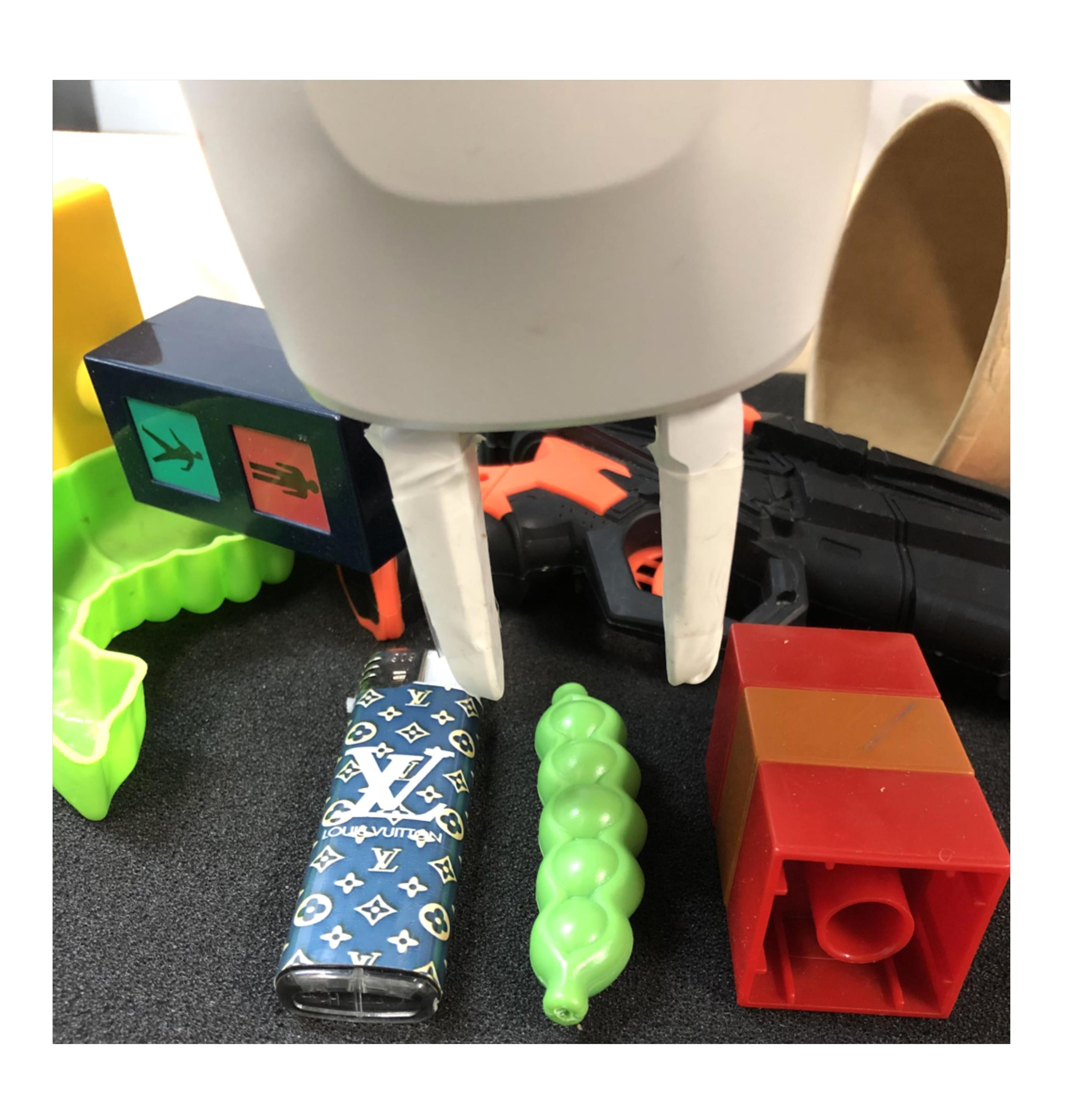}
         \label{fig:grasp_width_figure}}
    \caption{
     \subref{fig:grasp_width_distribution} Ground-truth and predicted grasp width distribution. \subref{fig:test_objects} An example shows adaptive grasp width in a clutter scene.}
     \vspace{-6mm}
\end{figure}

We first evaluate our proposed method in terms of grasp width. To demonstrate the high precision prediction of grasp width, we show a quantitative analysis of over 20k scene with around 2M synthetic grasps. In our experiments, we define the measurement for grasping width as the absolute difference between the predicted grasp width and the ground-truth grasp width $|\omega - \widehat{\omega}|$. We set 4 groups threshold for a comprehensive evaluation of grasp width prediction. For evaluation of each threshold, each absolute grasp width difference smaller than the threshold is classified as positive, otherwise negative. We select 100 proposals after NMS operation and filter out the negative samples. Experimental results shown in Tab.\ref{tab:clutter_grasp_width} demonstrate that our model can estimate high precision grasp width, and achieves 82.2\% accuracy under 5 mm threshold. Fig.\ref{fig:grasp_width_distribution} shows the overall grasp width distribution in our dataset. Grasp width is uniformly divided into 8 groups with an interval of 5 mm. While only 1/4 of all lie in the range $[3.5, 4]$ cm. Grasping with max opening width can be problematic in cluttered scenes, because it may lead to collisions with surrounding objects. Fig.\ref{fig:grasp_width_figure} shows an example that adaptive  grasp  width is critical for dexterous grasping in cluttered scenes.
\newcommand{\tabincell}[2]{\begin{tabular}{@{}#1@{}}#2\end{tabular}}
\begin{table}
\vspace{4mm}
    \centering
    \caption{Comparison of Grasp Width Accuracy}
        \begin{tabular}{c|c|c}
         \hline
        \multirow{2}*{\textbf{\tabincell{c}{Grasp Width\\Threshold(mm)}}} &\multicolumn{2}{c}{\textbf{Accuracy (\%)}} \\ \cline{2-3}
        &stage-1 & stage-2 \\ \hline
        2.5  & 42.0 & 52.5 \\
        5.0  & 76.0 & 82.2 \\
        7.5  & 87.5 & 90.2 \\
        10.0 & 92.9 & 93.1 \\ 
        \hline
    \end{tabular}
    \label{tab:clutter_grasp_width}
    \vspace{-6mm}
\end{table}

\subsubsection{One-stage VS. Two-stage}
To illustrate the effectiveness of our proposed grasp pose refinement network, we evaluate the generated grasp proposals quality for both the two stages.

As shown in Tab.\ref{tab:clutter_grasp_width}, grasp width accuracy after refinement has 25\% and 8\% improvement respectively over stage-1 under threshold 2.5 mm and 5 mm. The improvement gets saturated with higher tolerances. For grasp pose accuracy, we adopt the distance measurement of grasp pose as in Eq.\ref{equation:grasp_distance}. For evaluation of predicted grasp $g_p$, $g_p$ is classified as positive, when $D(g_p, g_t)$ is smaller than the predefined threshold, otherwise negative. Experimental results shown in Tab.\ref{tab:clutter_grasp_pose} demonstrate that proposals after refinement outperform stage-1 by a large margin. 
\vspace{-2mm}
\begin{table}
    \centering
      \caption{Comparison of Grasp Pose Accuracy}
      \begin{tabular}{c|c|c}
         \hline
        \multirow{2}*{\textbf{\tabincell{c}{Grasp Pose\\Threshold}}} &\multicolumn{2}{c}{\textbf{Accuracy (\%)}} \\ \cline{2-3}
        &stage-1 & stage-2 \\ \hline
        0.005  & 25.3 & 52.5 \\
        0.01  & 29.1 & 61.2 \\
        0.015 & 31.8 & 63.9 \\
        0.02  & 33.5 & 65.2 \\
        \hline
    \end{tabular}
    \label{tab:clutter_grasp_pose}
    \vspace{-2mm}
\end{table}
\subsection{Robotic Experiments}
\begin{figure}[t]
    \centering
    \subfigure[]{\includegraphics[width=0.35\linewidth]{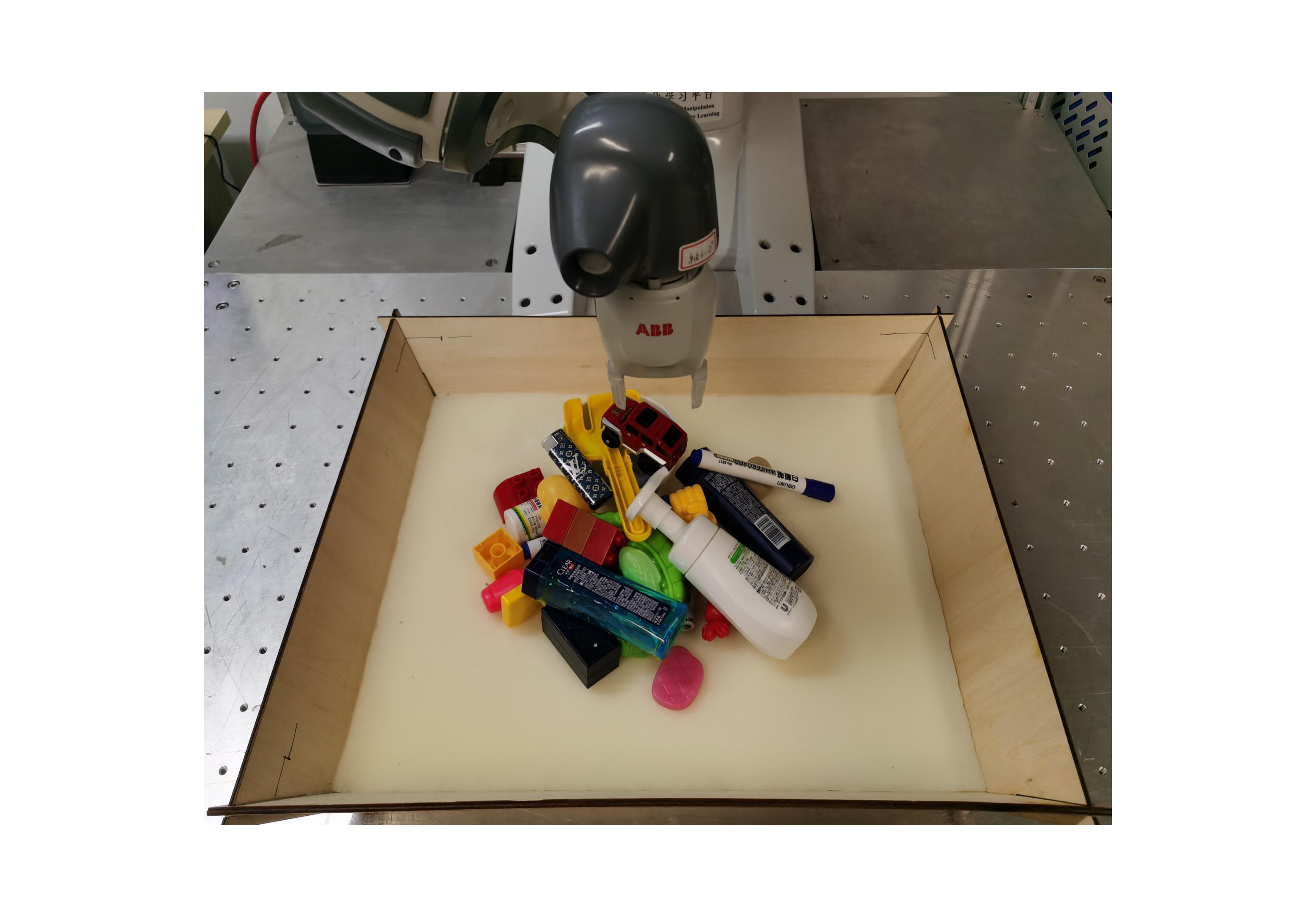}
        \label{fig:real_scene}}
    \subfigure[]{\includegraphics[width=0.46\linewidth]{figure/test_objects.pdf}
        \label{fig:test_objects}}
    \vspace{-2mm}
    \caption{Real setting of our robotic grasping experiments.
    \subref{fig:real_scene} Cluttered scene grasping experiment setup with ABB Yumi robotic arm. \subref{fig:test_objects} Objects used in our robotic experiments. Left one shows novel objects which are absent in the training dataset, right one shows similar objects.}
    \label{fig:real_scene_and_test_objects}
    \vspace{-6mm}
\end{figure}

We validate the reliability and efficiency of our proposed GPR network in ABB Yumi IRB-1400 robot and a PhoXi industrial sensor. Objects are presented to the robot in dense clutter as shown in Fig.\ref{fig:real_scene}. We keep a similar setting as in the simulation environment: 1) Camera is placed on top of the bin about 1.3 m; 2) Point cloud within the bin is cropped out for input data. 20 similar and 20 novel objects are selected for testing the generalization ability of our proposed network, as shown in Fig.\ref{fig:test_objects}. 

 We compare GPR to two state-of-the-art, open-sourced 6D grasp baselines, GPD \cite{gpd} and PointNetGPD \cite{pointnetgpd}. We train GPD and PointNetGPD with their default setting on our dataset with the code they released. 

The experiment procedure is as follows: 1) 10 of 20 objects are random sampled out, and then poured into the bin; 2) The robot attempts multiple grasps until all objects are grasped or 15 grasps have been attempted; 3)  10 times testing for each algorithm. The result is shown in Tab.\ref{tab:robot_experiments}. Success Rate (\textbf{SR}) and Completion Rate (\textbf{CR}) are used as the evaluation metrics.
\begin{table}[H]
    \vspace{-3mm}
    \centering
    \caption{Results of Clutter Removal Experiments}
    \begin{tabular}{c|cc|cc}
         \hline
         \multirow{2}*{\textbf{Method}} &\multicolumn{2}{c|}{\textbf{Similar objects}} &\multicolumn{2}{c}{\textbf{Novel objects}} \\
        \cline{2-5}
            & \textbf{SR}   & \textbf{CR} & \textbf{SR} & \textbf{CR}   \\
        \hline
        GPD (3 channels) \cite{gpd}    &  60\%  &  84\%  &  50\%   &  66\%    \\
        GPD (15 channels) \cite{gpd}   &  52.7\%  &  78\%  &  36\%  &   54\%    \\
        PointNetGPD (3 classes)\cite{pointnetgpd} &  64.6\%  &  84\%  &  54.8\%  &   80\%    \\
        \hline
        Ours& \textbf{78.3\%} & \textbf{94\%} & \textbf{69.2\%} & \textbf{90}\% \\
        \hline
    \end{tabular}
    \label{tab:robot_experiments}
    \vspace{-3mm}
\end{table}

As shown in Tab.\ref{tab:robot_experiments}, our method outperforms baseline methods in terms of Success Rate, Completion Rate, which demonstrates the superiority of our methods. In our observation, our algorithm can get better performance in terms of collisions with surrounding objects and stable grasp configuration.
\section{CONCLUSIONS}
In this paper, we proposed an end-to-end grasp pose refinement network for fine-tuning low-quality and filtering noisy grasps, which detects globally and refines locally. Meanwhile, we build a single-object grasp dataset which consists of 150 objects with various shapes, and a large-scale dataset for cluttered scenes. Experiments show that our model trained on the synthetic dataset performs well in real-world scenarios and achieves state-of-the-art performance.


\bibliographystyle{IEEEtran}
\bibliography{reference}

\end{document}